\documentclass[11pt]{article} 
\usepackage{rldmsubmit,palatino}
\usepackage{graphicx}
\usepackage{hyperref}

\title{Neuro-Nav: A Library for Neurally-Plausible \\ Reinforcement Learning}

\author{
Arthur Juliani \\
Araya Inc.\\
Minato-ku, Tokyo Japan\\
\texttt{arthur\_juliani@araya.org} \\
\And
Samuel A. Barnett \\
Princeton University \\
Princeton, NJ USA \\
\texttt{samuelab@princeton.edu} \\
\And
Brandon Davis \\
Massachusetts Institute of Technology \\
Cambridge, MA USA \\
\texttt{davisb@mit.edu} \\
\And
Margaret Sereno \\
University of Oregon \\
Eugene, OR USA \\
\texttt{msereno@uoregon.edu} \\
\And
Ida Momennejad \\
Microsoft Research \\
New York City, NY USA \\
\texttt{idamo@microsoft.com} \\
}

\begin{document}

\maketitle

\begin{abstract}

In this work we propose Neuro-Nav, an open-source library for neurally plausible reinforcement learning (RL). RL is among the most common modeling frameworks for studying decision making, learning, and navigation in biological organisms. In utilizing RL, cognitive scientists often handcraft environments and agents to meet the needs of their particular studies. On the other hand, artificial intelligence researchers often struggle to find benchmarks for neurally and biologically plausible representation and behavior (e.g., in decision making or navigation). In order to streamline this process across both fields with transparency and reproducibility, Neuro-Nav offers a set of standardized environments and RL algorithms drawn from canonical behavioral and neural studies in rodents and humans. We demonstrate that the toolkit replicates relevant findings from a number of studies across both cognitive science and RL literatures. We furthermore describe ways in which the library can be extended with novel algorithms (including deep RL) and environments to address future research needs of the field.
\end{abstract}


\startmain 

\section{Introduction}

Understanding how humans and other animals make non-reactive decisions in their environments remains a longstanding interest of psychologists and neuroscientists alike, especially following the proposal of cognitive maps a century ago \cite{tolman1948cognitive}. Applying the formalism of reinforcement learning (RL) to decision making has accelerated the progress in this field within recent decades \cite{sutton2018reinforcement}. Indeed, there is neuroscientific evidence that various forms of learning in humans and other animals can be modeled using the formalism of RL, and thus be described by classical algorithms from the RL literature \cite{niv2009reinforcement,momennejad2020learning}. 

Applying the RL paradigm has led to a deeper understanding of learning from describing simple forms of conditioning in animals \cite{sutton1990time}, to providing a basis for complex navigational strategies in humans \cite{simon2011neural}. This work is often made possible by abstracting the underlying complexity of the real environment-agent system into an idealized model. This process involves the real environment of the animal being abstracted into a computationally tractable Markov Decision Process (MDP) \cite{bellman1957markovian}, consisting of an underlying graph of state nodes, which are connected to one another via edges corresponding to actions taken by the agent. Likewise, the biological organism making the decisions is abstracted into an artificial agent modeled by a policy and value function, which can interact with and potentially learn the relevant dynamics of the MDP.

This process of abstraction is largely non-standardized, and has seen various ad-hoc implementations over the decades, often driven by the particular experimental needs of the researchers at the time. Furthermore, many of these concrete implementations are not available in open-source forms, change when translated from one coding language to another, or the code has been lost altogether. As a result, the ability to replicate or build on previous work has been limited, confusing trainees, slowing progress in the field, and preventing the straightforward comparison of various models or algorithms to animal data. On the other hand, most AI benchmarks remain limited to maximizing scores and super-human performance, which misses the opportunity for neurally and biologically plausible representation and behavior.

\begin{figure}[h!]
\includegraphics[width=18cm]{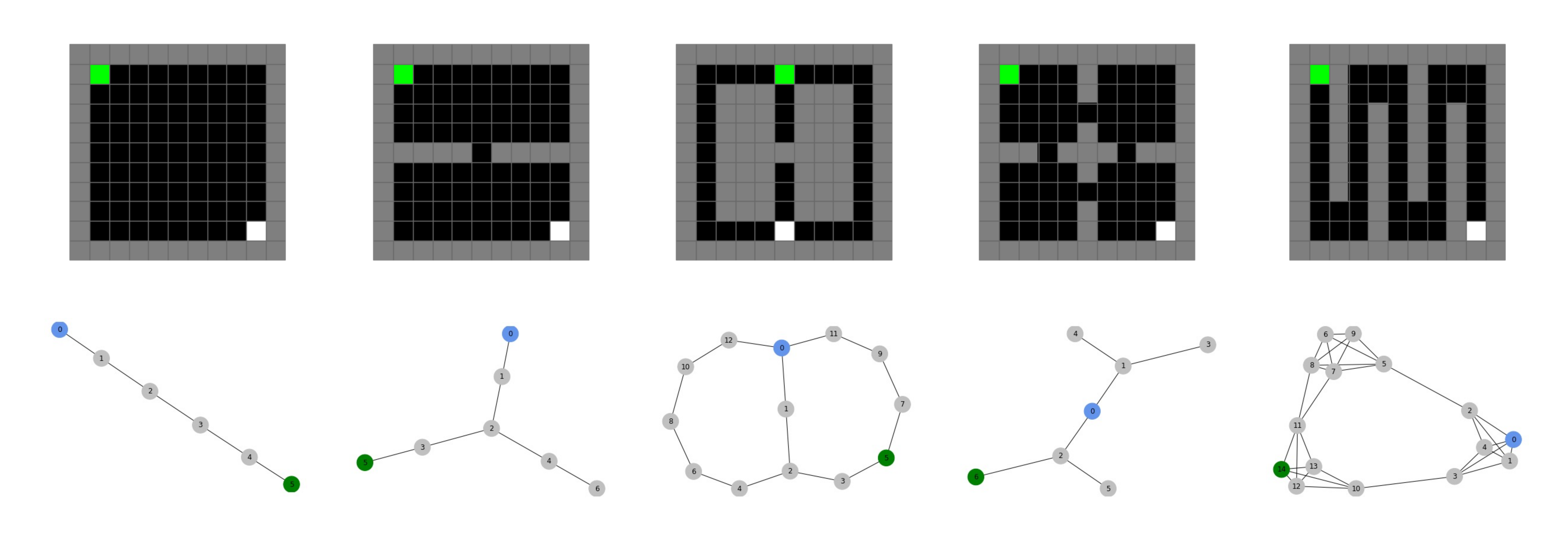}
\centering
\caption{\textbf{Top}: Examples of maze environments. White square corresponds to agent location. Green square corresponds to goal location. Grey squares correspond to walls.
\textbf{Bottom}: Examples of graph environments. Blue circle corresponds to agent location. Green circle corresponds to goal location. Environments drawn from studies in neuroscience \cite{schapiro2013neural,stachenfeld2017hippocampus}, classical RL \cite{sutton2018reinforcement}, and cognitive science \cite{momennejad2017human,russek2017predictive}. }
\label{example_fig}
\end{figure}

In this work, we present Neuro-Nav\footnote{\href{https://github.com/awjuliani/neuro-nav}{https://github.com/awjuliani/neuro-nav}}, an open-source library providing a standardized set of benchmark environments and RL algorithms which can be used to both replicate previous findings, test the biological plausibility of new models, and provide scaffolding for future experimental work in the field. The benchmarks focus on domains that are the basis of many experimental studies: spatial navigation in mazes (Figure \ref{example_fig} Top), associative learning, and graph navigation tasks (Figure \ref{example_fig} Bottom) \cite{momennejad2020learning}. In addition, we provide standardized implementations of classical RL algorithms such as Q-Learning, Successor Representation, and Model-based RL with Value Iteration, which can be evaluated using the benchmark environments. We provide these all within the context of a documented and tested open-source repository, which can be freely used and developed further by the broader research community. 

\section{Neuro-Nav Library}

Neuro-Nav is an open source library consisting of three components: a set of benchmark environments, a toolkit containing artificial reinforcement learning agents and algorithms, and a set of interactive notebooks for replicating experimental findings in the literature. All of the code in Neuro-Nav is written in the python programming language, and utilizes the NumPy and PyTorch computational programming libraries to accelerate the relevant tensor mathematics.

\subsection{Benchmark Environments}

The environments included in Neuro-Nav consist of a set of graph and maze navigation tasks. In both cases, the underlying environment dynamics consists of a graph of `state' nodes connected by `action' edges. The agent occupies one of the nodes in the graph. The graph may also contain nodes that provide reward upon the agent occupying them, as well as a goal node, which both provides a reward as well as terminates the episode. Maze environments may be seen as a specific class of graph environments where the structure is mapped to a 2D grid, and the action space always consists of four actions: movement in each of the cardinal directions. Both types of environments are implemented using the OpenAI Gym interface, thus allowing for integration with a wide number of pre-existing open source RL codebases \cite{brockman2016openai}.

For each class of environments, we provide both an underlying library for defining and interacting with the environment as well as a set of pre-defined environments, which can be utilized for evaluation. These environments are often taken from existing literature, and serve to aid in the partial or complete replication of various findings from those original papers. In particular, we draw from environments defined in the context of neuroscience \cite{schapiro2013neural,stachenfeld2017hippocampus}, classical RL \cite{sutton2018reinforcement}, and cognitive science \cite{momennejad2017human,russek2017predictive}. See Figure \ref{example_fig} for examples of a representative subset of the included maze and graph environments.

We also provide an abstraction for defining the observation space of the environment, which is completely orthogonal to that of the structure or topography of the environment. By default, we utilize a one-hot encoding for states, but also provide a set of alternative observational spaces such as Euclidean distance from walls in the maze tasks, or CIFAR images in the graph tasks, as well as many others. 

\subsection{Algorithms Toolkit}

\begin{table}[h]
\centering
\begin{tabular}{||c c c c||} 
 \hline
 Algorithm & Function(s) & Update Rule(s) & Reference \\ [0.5ex] 
 \hline\hline
 TD-Q & $Q(s, a)$ & one-step temporal difference & \cite{watkins1992q} \\ 
 \hline
 TD-SR & $\psi(s, a)$, $\omega(s)$ & one-step temporal difference & \cite{dayan1993improving} \\
 \hline
 TD-AC & $V(s)$, $\pi(a | s)$ & one-step temporal difference & \cite{sutton2018reinforcement} \\
 \hline
 Dyna-Q & $Q(s, a)$ & one-step temporal difference, replay-based dyna & \cite{sutton1990integrated} \\
 \hline
 Dyna-SR & $\psi(s, a)$, $\omega(s)$ & one-step temporal difference, replay-base dyna & \cite{russek2017predictive} \\
 \hline
 Dyna-AC & $V(s)$, $\pi(a | s)$ & one-step temporal difference, replay-based dyna & N/A \\
 \hline
 MBV & $Q(s, a)$, $T(s' | s, a)$ & value-iteration & \cite{sutton2018reinforcement} \\
 \hline
 MBSR & $\psi(s, a)$, $\omega(s)$, $Q(s, a)$, $T(s' | s, a)$ & value-iteration, one-step temporal difference & \cite{momennejad2017human} \\
 \hline
 QET & $Q(s, a)$, $e(s, a)$ & eligibility trace & \cite{sutton2018reinforcement} \\
 \hline
\end{tabular}
\caption{Reinforcement learning algorithms included in the Neuro-Nav toolkit.}
\label{agent_table}
\end{table}

In addition to graph and maze environments, we provide a set of artificial agents, which implement canonical algorithms from the RL literature. We focus on tabular rather than deep learning algorithms, as this is often the focus of the literature at the intersection of neuroscience and RL \cite{niv2009reinforcement,russek2017predictive}. We include algorithms implemented either as model-free or model-based learning algorithms, as well as hybrid algorithms such as Dyna. See Table \ref{agent_table} for a description of the provided agent types. Each agent implements one of multiple possible algorithms for learning a policy and value function. Decisions are taken by the agent with either a softmax policy or epsilon greedy strategy, both of which are available to use in any agent type.

\section{Experimental Results}

\begin{figure}[h]
\includegraphics[width=18cm]{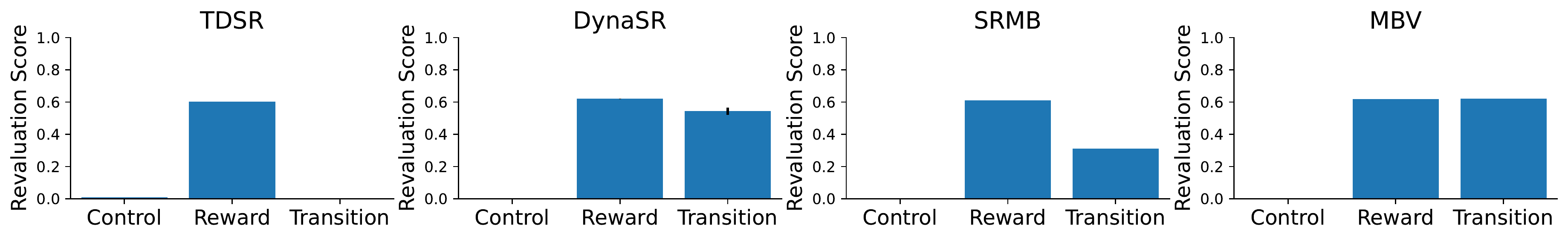}
\centering
\caption{Replication of results from revaluation task described in Experiment 1 from \cite{momennejad2017human}. Higher revaluation score corresponds to greater change in behavior after the re-learning phase. Both DynaSR and SRMB algorithms show most human-like revaluation behavior, which displays greater revaluation for reward changes compared to transition changes. Results averaged over ten experiments.}
\label{graph_1_res}
\end{figure}

We validate Neuro-Nav by replicating known results from human and rodent navigation and decision-making experiments. In particular, we replicate two classes of results: behavioral, which compares the observable behavior of biological and artificial agents, and representational, comparing learned representations of biological agents (using neuroscience) and artificial agents. Here we present Neuro-Nav results using benchmark from both categories.

\begin{figure}[h]
\includegraphics[width=18cm]{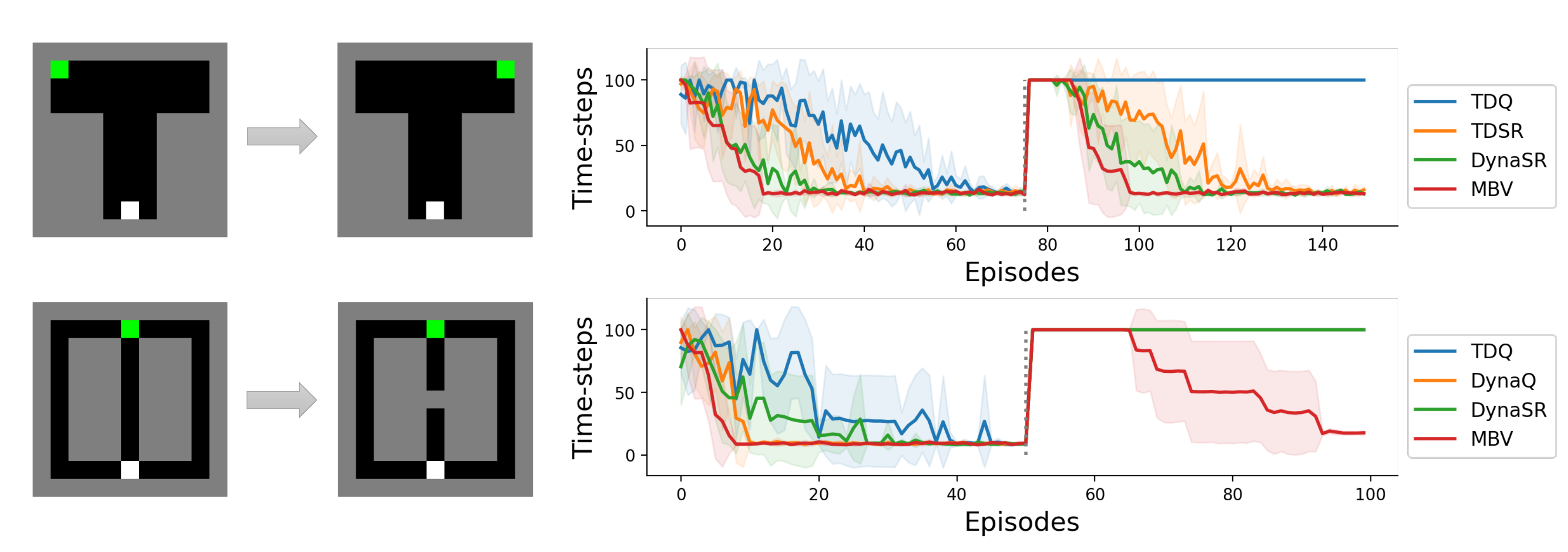}
\centering
\caption{\textbf{Top}: Performance comparison between agent types on a reward transfer problem. Goal location changes at episode 75. SR and model-based algorithms successfully adapt to change. \textbf{Bottom}: Performance comparison between agents on a structure transfer problem. Environment structure changes at episode 50. Only model-based algorithm adapts to structural change. Graphs display average performance over five separate experiments. Tasks adapted from \cite{russek2017predictive}.}
\label{maze_res}
\end{figure}

In order to validate Neuro-Nav for replication of behavioral results, we focused on two recent studies in the literature, \cite{momennejad2017human} and \cite{russek2017predictive}. In the former, agent performance is compared to that of humans in a set of navigation tasks, where aspects of the environments are changed partway during learning. In the latter, agent performance is compared to that of rodents in a set of maze navigation tasks. We present replications of the findings from Experiment 1 of \cite{momennejad2017human} in Figure \ref{graph_1_res}. We present results consistent with the results of \cite{russek2017predictive} in Figure \ref{maze_res}.

\begin{figure}[h!]
\includegraphics[width=12cm]{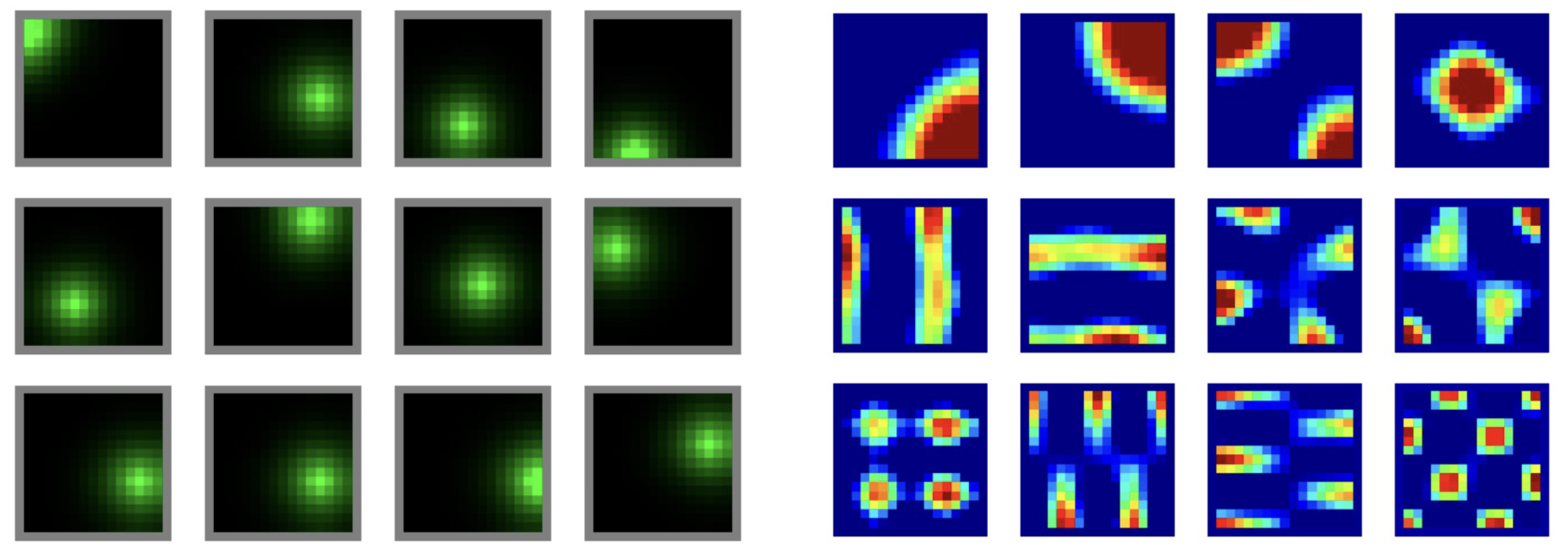}
\centering
\caption{\textbf{Left}: Values of a selection of units in the $\psi(s)$ function in a maze environment after learning. Units show place-like spatially selective fields. \textbf{Right}: PCA of $\phi(s)$ function in a maze environment after learning. Units show grid-like spatial selectivity. Procedure for generating place and grid fields adapted from \cite{stachenfeld2017hippocampus}.}
\label{reps_fig}
\end{figure}

\begin{figure}[h!]
\includegraphics[width=12cm]{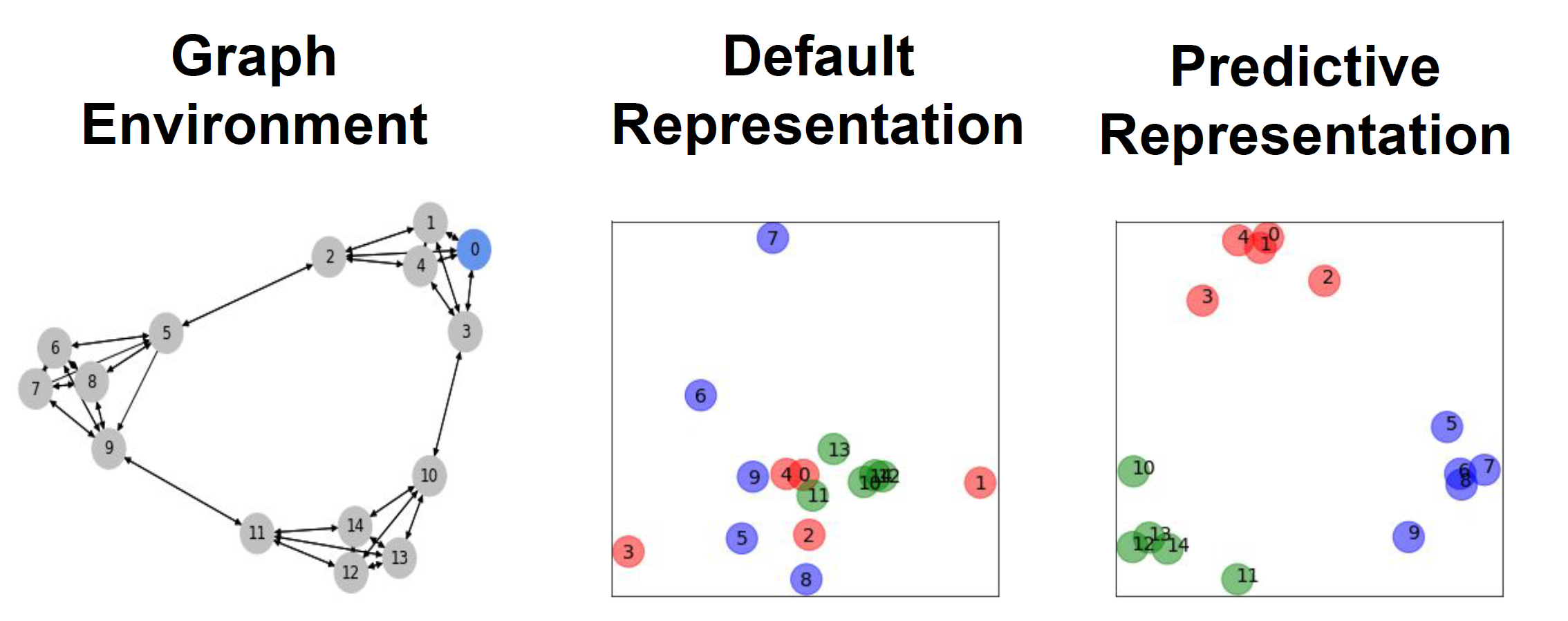}
\centering
\caption{Latent representation of model trained with predictive representation shows temporal community structure. \textbf{Left:} Graph environment with three distinct neighborhoods of state nodes. \textbf{Middle:} Dimensionality reduction of default one-hot state encoding. Colors represent neighborhoods. \textbf{Right:} Dimensionality reduction of latent representation of one-hidden-layer neural network trained to predict next state occupancy. Temporally similar states are represented similarly, as is the case in various brain regions \cite{schapiro2013neural}.}
\label{tcs_fig}
\end{figure}

Beyond simply analyzing behavior, the internal content and structure of information utilized for decision making is of interest to many researchers. Here we focus on replicating the main computations results of two different experiments. The first is the emergence of place and grid like cells from a successor representation \cite{stachenfeld2017hippocampus}, which can be seen in Figure \ref{reps_fig}. The second is the emergence of temporal community structure, a computational property found in various brain regions, from a neural network trained with a predictive task \cite{schapiro2013neural}, which can be seen in Figure \ref{tcs_fig}.

\section{Discussion}

In this work we presented an open-source library of benchmark environments and algorithms for performing reinforcement learning experiments on decision making and navigation tasks. Neuro-Nav aims to empower reproducibility and standardization of evaluation within research in neurally plausible RL models of navigation and decision making. This project was developed with future extensions in mind, and as such, Neuro-Nav empowers users to easily develop new environments and algorithms beyond what is demonstrated in this work. 

Neuro-Nav users can extend the benchmark environments in two ways. The first is by creating novel environments, i.e., MDPs, graphs, and maze topographies, to evaluate specific navigation, associative learning, or decision making experiments. The second is by defining novel observation spaces for the agents, such as using fractal or face/scene images as nodes or states, or perceiving Euclidean distance from maze walls. These examples, among others, capture relevant paradigms in the neuroscience literature \cite{schapiro2013neural,de2020neurobiological}. Future work using Neuro-Nav can add novel environments and novel algorithms to the open-source library, compare the performance of RL algorithms on all environments, and thus replicate a broader class of studies within the field.

Currently, the Neuro-Nav library only supports tabular learning agents. As such, the agents provided here are not capable of learning from the more varied class of possible observation spaces. In contrast, humans and other animals learn from complex multi-dimensional sensory signals. While much unrelated work in the field of deep RL has focused on learning from high-dimensional observation spaces \cite{arulkumaran2017deep}, we believe there is an opportunity for a middle path of lower-dimensional, but biologically grounded observation spaces, which enables linear models or simple neural networks to learn more expressive and generalizable behavioral policies. While we plan to include deep RL algorithms in future versions, we believe the current version of Neuro-Nav offers a promising step toward biologically plausible benchmarks, and a toolkit with potentially significant contributions to the field. 

\bibliographystyle{unsrt}

\bibliography{references.bib}

\end{document}